\documentclass{article}

\usepackage{arxiv}
\usepackage{natbib}
\usepackage[utf8]{inputenc} 
\usepackage[T1]{fontenc}    
\usepackage{hyperref}       
\usepackage{url}            
\usepackage{booktabs}       
\usepackage{amsfonts}       
\usepackage{nicefrac}       
\usepackage{microtype}      
\usepackage{lipsum}
\usepackage{graphicx}
\usepackage{algorithm,algpseudocode}
\usepackage{bm}
\usepackage{makecell}
\usepackage{booktabs}
\usepackage{stmaryrd}
\usepackage{graphicx}
\usepackage{threeparttable}
\usepackage{subcaption}
\usepackage{amsmath}

\newtheorem{definition}{Definition}%
\graphicspath{ {./images/} }

\title{Inferring Preferences from Demonstrations in Multi-objective Reinforcement Learning \thanks{This work has been published in Neural Computing \& Application (2024). \href{https://doi.org/10.1007/s00521-024-10412-x}{https://doi.org/10.1007/s00521-024-10412-x}}}

\author{
 Junlin Lu \\
  School of Computer Science\\
  University of Galway\\
  University Road, Galway, H91 TK33, Ireland \\
  \texttt{j.lu5@universityofgalway.ie} \\
   \And
 Patrick Mannion \\
  School of Computer Science\\
  University of Galway\\
  University Road, Galway, H91 TK33, Ireland \\
  \texttt{patrick.mannion@universityofgalway.ie} \\
  \And
 Karl Mason \\
  School of Computer Science\\
  University of Galway\\
  University Road, Galway, H91 TK33, Ireland \\
  \texttt{karl.mason@universityofgalway.ie} \\
}

\begin{document}
\maketitle
\begin{abstract}
Many decision-making problems feature multiple objectives where it is not always possible to know the preferences of a human or agent decision-maker for different objectives. However, demonstrated behaviors from the decision-maker are often available. This research proposes a dynamic weight-based preference inference (DWPI) algorithm that can infer the preferences of agents acting in multi-objective decision-making problems from demonstrations. The proposed algorithm is evaluated on three multi-objective Markov decision processes: Deep Sea Treasure, Traffic, and Item Gathering, and is compared to two existing preference inference algorithms. Empirical results demonstrate significant improvements compared to the baseline algorithms, in terms of both time efficiency and inference accuracy. The DWPI algorithm maintains its performance when inferring preferences for sub-optimal demonstrations. Moreover, the DWPI algorithm does not necessitate any interactions with the user during inference - only demonstrations are required. We provide a correctness proof and complexity analysis of the algorithm and statistically evaluate the performance under different representation of demonstrations.
\end{abstract}


\section{Introduction}\label{sec1}

Many decision-making tasks are designed with the primary goal of optimizing performance in achieving a singular objective, such as enhancing game scores \cite{mnih2015human}, reducing energy consumption \cite{lu2023go}, and ensuring appropriate robotic functionality \cite{kober2013reinforcement}, among others. Nonetheless, in practical scenarios, more decision-making tasks feature multiple objectives. This complexity introduces the necessity for a balanced trade-off among these objectives. Given the variability in user preferences regarding objectives, the introduction of a preference factor, typically represented by a linear weight vector \cite{hayes2022practical}, is essential for the comparative evaluation of distinct solutions.

Multi-objective reinforcement learning (MORL) is the paradigm to solve such multi-objective decision-making (MODM) problems, by considering the preference factor. If the preference factor is known beforehand, a standard policy improvement approach can be used\cite{mannor2001steering, tesauro2007managing, van2013scalarized}. This, however, cannot work properly without knowing the preference beforehand and it needs to be re-trained from scratch when the preference changes. Though there are more advanced works that can work with dynamic preferences \cite{kallstrom2019tunable, yang2019generalized, alegre2023sample}, it cannot give the correct solution when the user cannot say out true preference. It is challenging for the user to numerically specify the preference corresponding to their true thoughts. For example, when selecting stocks for a portfolio, the portfolio manager selects stocks based on their weighting of minimizing risk and maximizing potential future profits. He/she might want to give a higher weighting to maximize potential future profits, but should the weight be 0.7, 0.8, or some other number? Setting weights this way is unintuitive and must typically be done by trial and error. Moreover, even if the user can provide an approximate numerical preference, a small error in their preference can result in a significantly different policy being learned and executed which may lead to a sub-optimal solution. 

While it may be challenging for users to express preferences, their behavior/demonstration can provide insights into their true thoughts. The process of inferring preference from demonstraions is known as preference inference (PI) \cite{lu2023inferring,lu2023preference,lu2024inferring,hwang2023promptable}. By leveraging user demonstration to infer their true thoughts, we can make more effective and user-friendly MODM systems.

Although there is some previous work on the PI for MODM, inferring preferences from demonstrations is still challenging and has yet to receive enough attention in the literature. Many PI methods involve active learning approaches, which require feedback from users \cite{benabbou2015incremental,zintgraf2018ordered,benabbou2020interactive,shao2023eliciting}. Other methods use ideas from inverse reinforcement learning (IRL) \cite{ng2000algorithms,ziebart2008maximum} to infer preferences from demonstrations and heuristically search for the correct weight vector in the preference space \cite{ikenaga2018inverse,takayama2022multi}, which can be computationally expensive and suffer from low accuracy without optimal demonstrations. MORL approaches that learn the parameterized policy set can also be used to infer user preference \cite{barrett2008learning, yang2019generalized}. These PI paradigms either suffer from high computing overload, frequent interaction with the user, or brittleness when sub-optimal demonstrations are given. It would therefore be advantageous to develop a new method that can perform autonomous query-free PI at a low computational cost and is robust against sub-optimal demonstrations.

We propose a dynamic weight-based preference inference (DWPI) algorithm that aims to train a PI model that is free of querying user and robust against sub-optimal demonstrations. Our method does not need the user to provide any training data. The training data is all generated by a dynamic weight MORL (DWMORL) agent. We evaluate the proposed algorithm in three different environments featured with various number of objectives, namely Convex Deep Sea Treasure (CDST) \cite{mannion2017policy}, Traffic, and Item Gathering \cite{kallstrom2019tunable}. The DWPI algorithm involves training a deep neural network model for PI using a feature set of demonstration and a target set of preference weight vectors generated by the DWMORL agent in simulations. Compared to existing approaches, the DWPI algorithm avoids the computational overhead and requires no user feedback than PI paradigms with active learning. Additionally, the DWPI algorithm is more robust than other methods in the literature as it can infer preferences correctly even if the demonstration is stochastic and sub-optimal. The main contributions of this paper are:

\begin{itemize}
    \item We propose the DWPI algorithm, a time-efficient, robust, query-free, and high-accuracy method for inferring preferences from demonstrations (including sub-optimal demonstrations) in MORL settings. After the model is trained, the model only requires a forward pass to infer the preference for a given demonstration.
    \item The DWPI algorithm's generalization capability is assessed across three distinct environments, each featuring a varying number of objectives, to demonstrate its adaptability and effectiveness.
    \item We provide a proof of correctness for the utilization of a feedforward neural network (FNN) in the PI process from demonstrations. Additionally, we perform a complexity analysis of our algorithm and introduce a metric to evaluate the accuracy of the inference process.
    \item An energy-based model is introduced to deliberately generate sub-optimal demonstrations, thereby enriching the training dataset for the PI model, and enhancing its learning potential.
    \item The performance of the DWPI algorithm is evaluated under various demonstration representations, underscoring its versatility and efficiency in processing diverse data inputs.
\end{itemize} 

The remainder of this paper is organized as follows. We outline the necessary background knowledge for this paper in Section 2. In Section 3 we present the formal model for the DWPI algorithm. In Section 4, we illustrate the design of our experiment The results of the experiments are demonstrated and discussed in Section 5. In Section 6, we conclude the paper and propose some future research directions that have arisen from this work. 

\section{Background \& Related Work}
\subsection{Multi-objective Reinforcement Learning}
Multi-objective reinforcement learning (MORL) is a branch of reinforcement learning that focuses on decision-making problems with multiple objectives \cite{hayes2022practical, van2013scalarized, van2014multi, mossalam2016multi}. In MORL, an agent interacts with the environment and receives a reward vector composed of multiple objective values, rather than a single scalar reward like in single-objective reinforcement learning (SORL). However, this increases the computation complexity of training. To make training more manageable, reward scalarization is often used in MORL. This involves mapping the reward vector to a scalar value using a scalarization function, also known as a utility function. The utility function $u:\mathbb{R}^{d}\xrightarrow{}\mathbb{R}$ is given below,
\begin{equation}
    V_{u}^{\pi}=u(\bm{V}^{\pi})
\end{equation}
where the $\bm{V}^{\pi}$ is the multi-objective value vector when adopting policy $\pi$ and $V_{u}^{\pi}$ is the scalarized value of policy $\pi$. 

Linear scalarization is a commonly used method in MORL and has been applied in various studies, where weights are assigned to each of the objectives to compute a weighted sum \cite{abels2019dynamic,barrett2008learning,castelletti2012tree,roijers2017interactive,van2013scalarized}. The weights reflect the relative importance of the different objectives. The resulting scalar value (utility) can be used to compare different policies.

The convex coverage set (CCS($\Pi$)) is a subset of the $\Pi$, where $\Pi$ is the set of all possible policies. CCS($\Pi$) contains optimal policies for every linear weight vector \cite{hayes2022practical}. Within the scope of using linear utility function, CCS($\Pi$) contains all necessary information for optimal policy set, i.e. the CCS($\Pi$) represents the set of Pareto optimal solutions that are guaranteed optimal for any linear combination of the objectives.

\subsection{Preference Inference}
\label{subsec: preference inference}
Previous studies have mainly focused on uncovering the user's unknown preferences within MODM?. These methodologies are broadly categorized into approaches that leverage queries to elicit preferences \cite{benabbou2015incremental,zintgraf2018ordered,benabbou2020interactive,shao2023eliciting},  inverse reinforcement learning paradigm (IRL) \cite{ikenaga2018inverse,takayama2022multi}, and using a learned policy set to do the inference \cite{barrett2008learning, yang2019generalized}. We start with the query-based methods.

Incremental Weight Elicitation (IWE): Benabbou et al. proposed a method utilizing a vector-valued state-space graph \cite{benabbou2015incremental}. This approach begins by identifying a set of potential preferences, followed by employing incremental queries to narrow down the volume of candidate preferences until an approximate optimal solution is identified.

Active Learning Method - Ordered Preference Elicitation (OPE): Zintgraf et al. introduced an active learning strategy termed as ordered preference elicitation \cite{zintgraf2018ordered}. This method presupposes the obscurity of the user's intrinsic preference and adopts a Gaussian process to model this latent preference. It features three querying strategies: pair comparison, ranking, and clustering, aimed at aligning the policy with the user's underlying preferences.

Regret-Based Incremental Preference Elicitation (RIGA): Employing a genetic algorithm, this approach, as discussed by Benabbou et al. \cite{benabbou2020interactive}, also utilizes incremental querying. However, unlike the IWE method, RIGA applies incremental queries to select instances from a set rather than to diminish the set of candidate preferences.

Comparative Feedback/Query (CF) for Eliciting Preference: Recent work by Shao et al. \cite{shao2023eliciting} advances the elicitation process by soliciting user feedback on pairs of policies or through weighted representative demonstration feedback. This method facilitates preference elicitation by enabling direct comparison between different policy outcomes.

It is not explicitly stated in the original contributions whether these query-based methods work for sub-optimal demonstrations. However, they might be able to find out the underlying preferences even from sub-optimal demonstrations as the query involves the user iteratively updating the solution.

Another branch is to use the IRL paradigm. Ikenaga et al. \cite{ikenaga2018inverse} proposed a PI approach, i.e., projection method (PM) based on the inverse reinforcement learning (IRL) paradigm. Their model starts from a heuristic preference and compares the expected feature (e.g. cumulative rewards) of an expert policy, and the expected feature obtained using the inferred preference. The distance between the ground truth preference and the inferred preference is measured. The inference process is considered correct once the reward expectation of the inferred preference converges to the expectation of the ground truth preference. However, the inferred preference is from random sampling from the preference space, and at each round of inference, the RL agent must be trained from scratch each time to obtain the reward expectation of the inferred preference. This can be time-consuming for high-dimensional preference spaces. To address this, Takayama et al. \cite{takayama2022multi} proposed a more efficient method based on deep reinforcement learning (DRL) and multiplicative weights apprenticeship learning (MWAL). Rather than randomly sampling, this method updates the inference along the direction indicated by the difference between the feature expectation of the expert policy and the feature expectation of the inferred preference. However, training an RL agent from scratch is still one of the bottlenecks for this approach. Their approaches have a further limitation in that they do not take into account sub-optimal demonstrations, where the demonstrator may not possess complete expertise for the problem at hand. 

Barrett et al. have developed convex hull value iteration (CHVI) \cite{barrett2008learning}. This method primarily focuses on learning to solve MORL tasks, with the PI task considered an auxiliary application. The CHVI method facilitates PI through a systematic comparison between the agent's action selections and the acquired knowledge of Q-values, enabling the identification of the preference weight set that aligns with the agent's decisions. By applying intersection operations on these weight constraints, the method arrives at a refined inference of the preference vector.

The envelope MOQ-learning algorithm (EMOQL) proposed by Yang et al. introduces a mechanism whereby the agent learns an optimal policy that accommodates all conceivable preferences. Their research posits that EMOQL is capable of elucidating the underlying preference vector. The assumption is that there exists a latent preference factor capable of converting the reward vector to a scalar, $r$. By maintaining a constant policy set and traversing the preference space, it is possible to determine which preference maximizes the cumulative reward scalar with signals when scalarized by this hidden preference \cite{yang2019generalized}. The identified preference is thus considered the most accurate representation of the latent preference factor.

In conducting a comparative analysis of PI methods, as summarized in Table \ref{tab:Comparative Analysis Table of Preference Inference Methods}, two key criteria are employed for evaluation: the dependency on querying to elicit preference, and the capability to handle sub-optimal demonstrations.

Specifically, it distinguishes between methods that require active engagement with users to refine PIs through queries and those that autonomously infer preferences from observed behaviors. Additionally, it assesses the resilience of these methods in deriving accurate preference insights from demonstrations that may not represent the optimal policy execution. This highlights the adaptability to real-world complexities where ideal demonstration data may be unavailable. This comparative framework facilitates an understanding of the strengths and limitations inherent to each PI approach.

\begin{table}[ht]
\centering
\caption{Comparative Analysis}
\label{tab:Comparative Analysis Table of Preference Inference Methods}
\begin{tabular}{cccc}
\toprule
Method & Free of Querying Users & Robust to Sub-optimal Demos & Reference\\
\midrule
IWE& $\times$ & $\checkmark$ &\cite{benabbou2015incremental}\\
\midrule
OPE& $\times$  & $\checkmark$ & \cite{zintgraf2018ordered}\\ 
\midrule
RIGA& $\times$  & $\checkmark$ &\cite{benabbou2020interactive}\\ 
\midrule
CF& $\times$  & $\checkmark$ & \cite{shao2023eliciting}\\ 
\midrule
PM& $\checkmark$  & $\times$ & \cite{ikenaga2018inverse}\\
\midrule
MWAL& $\checkmark$  & $\times$ &\cite{takayama2022multi}\\
\midrule
CHVI& $\checkmark$  & $\times$ & \cite{barrett2008learning}\\
\midrule
EMOQL & $\checkmark$  & $\times$ & \cite{yang2019generalized}\\
\midrule
DWPI (ours)& $\checkmark$ & $\checkmark$ & -\\
\bottomrule
\end{tabular}
\end{table}

\subsection{Problem Statement}
\label{Problem Statement}
As highlighted, the deployment of MORL agents in real-world decision-making necessitates an understanding of the user's genuine preferences regarding the weighting attributed to various objectives. Regrettably, the articulation of numerical preferences, which typically manifest as a linear weight vector \cite{castelletti2013multiobjective,khamis2014adaptive,ferreira2017multi,lu2022multi}, is not inherently intuitive for the user. In adverse scenarios, even when a user is capable of specifying a distinct preference, this preference must be sufficiently precise, as minor discrepancies can lead to the adoption of an inappropriate policy. Such obstacles hinder the application of MORL methods in practice. This motivates the study in PI methods for MORL scenario.

Though there are PI methods in the literature, as we mentioned in Section \ref{subsec: preference inference}, these approaches often require querying the user, face challenges with computational complexity in scenarios involving high-dimensional preference spaces, or struggle to accommodate sub-optimal demonstrations.

Our target is to introduce a PI model designed to surmount the previously mentioned limitations. We employ a dynamic weight multi-objective reinforcement learning (DWMORL) agent \cite{kallstrom2019tunable} to produce data for the inference model. The advantage of self-generating training data is the elimination of the need for user queries during the training phase. Once the DWMORL agent has been adequately trained, the inference procedure becomes a straightforward feedforward operation, irrespective of the number of objectives involved. To facilitate the inference of preferences from sub-optimal demonstrations, we enable the DWMORL agent to generate sub-optimal data. This approach allows for the augmentation of the training set with sub-optimal demonstrations. We assess our methodology across three MORL benchmarks: CDST \cite{mannion2017policy}, Traffic, and Item Gathering \cite{kallstrom2019tunable}.

We wish to clarify that the term ``preference". The term used in our discourse is specifically defined within the context of MORL, distinguishing it from its usage in discussions surrounding the integration of IRL and SORL. In the work of Hejna et al. \cite{hejna2023few}, they mentioned ``preference” as well, however, it is used to rank the demonstration of the robot, to navigate it towards a better policy. It is different from the ``preference” in our work where it is a linear weight vector to balance between several known objectives. 
The target of this method is also different from ours. 
They focus on involving human feedback to improve the task performance while we aim at inferring the users’ feature of preference which can be used for opponent modelling etc.  

\section{DWPI Algorithm}
In this section, we commence by presenting the DWMORL agent, which can adjust to changing priorities across multiple objectives. It plays a pivotal role in producing the training dataset for the DWPI model. Subsequently, we introduce the concept of truncated Q soft Q action sampling, which is employed to create sub-optimal demonstrations to enhance the training dataset further. Following this, we elucidate the DWPI algorithm. The section continues with a theoretical analysis that encompasses both a proof of correctness and an analysis of complexity, and the bi-directional mapping between demonstrations and preferences. In Table \ref{tab:symbols} We put a summary of the symbols used in this section.

\begin{table}[ht]
\centering
\begin{tabular}{cc|cc}
\toprule
Symbol & Meaning &Symbol & Meaning\\
\midrule
$Env$&environment & $\alpha$ & learning rate\\
$\gamma$&discount factor&$\epsilon$&exploration factor\\
$\pi_{\epsilon}$&policy conditioned on $\epsilon$ &$M$&number of episodes\\
$H$&horizon length&$\bm{w}$&preference weight vector\\
$\mathcal{W}$&preference weight simplex&$\eta$&the granularity discretize $\mathcal{W}$\\
$\mathcal{W}^{\eta}$&discretized subset of &$Q_{\bm{w}}$&Q table labelled by $\bm{w}$\\
$a_{t}$&the action sampled at timestep $t$&$s_{t}$&the state at timestep $t$\\
$\bm{r}_{t}$&the reward vector at timestep $t$&$r_{t}$&the reward scalar at timestep $t$\\
$\delta_{t}$&the TD-error at timestep $t$&$\pi_{soft}^{+}$ & truncated soft policy\\
$\pi^{*}$&the optimal policy&$\delta$ & policy equivalence threshold\\
$\theta$&the Q network&$\theta_{-}$&the target Q network\\
$\pi_{\epsilon,\theta,\bm{w}}$ &policy conditioned on $\epsilon,\theta,\bm{w}$&$\mathcal{D}$& replay memory\\
$N_{min}$ & volume threshold of $\mathcal{D}$&$\phi$& the PI model\\
$\pi$ & DWMORL agent(policy) &$\tau_{\bm{w}_{i}}$& demonstration based on $\bm{w}_{i}$\\
$\mathcal{L}$ & loss function&$\nabla$& gradient operator\\
$\beta$ & Boltzmann distribution temperature&$N$& number of actions retained\\
$a$ & augmented factor for training data&$E$& DWPI train epoch number\\
$\bm{R}^{n}$ & $n$-dimensional reward space&$n_{l}$& $l$th layer neuron number\\
\bottomrule
\end{tabular}
\caption{Notation and Nomenclature of Algorithmic Terms}
\label{tab:symbols}
\end{table}


\subsection{Dynamic Weights MORL Agent Training}
\label{subsec: Train Decoder}
We employed the methodology proposed by Kallström et al. to train the DWMORL agent \cite{kallstrom2019tunable}. Specifically, we adpated this approach to implement both tabular and DRL Q-agents, which are respectively referred to as the DWMOTQ (dynamic weights tabular Q-learning) and DWMODQN (dynamic weights deep Q-network) algorithms. We partitioned the preference weight space $\mathcal{W}$ into a discrete subset $\mathcal{W}^{\eta}$ using a specific granularity $\eta$ (e.g., $\eta=0.01$ for DST), such that $\mathcal{W}^{\eta} \subseteq \mathcal{W}$.

\subsubsection{Dynamic Weights Multi-objective Tabular Q-learning}
The DWMOTQ algorithm is outlined in Algorithm \ref{alg: Q Learning with tunable training}. The DWMOTQ agent keeps a collection of Q tables, each indexed by a preference weight vector, to maintain a record of action-state values. 

At the beginning of each episode, a preference weight $\bm{w}$ is randomly sampled from $\mathcal{W}^{\eta}$. The DWMOTQ agent then interacts with the environment and updates the Q table associated with $\bm{w}$ until it converges.
\begin{algorithm}[ht]\small
\caption{DWMOTQ Algorithm}
\label{alg: Q Learning with tunable training}
\begin{algorithmic}
\State {\textbf{Initialize:}}
\State {$Env$ (environment), $\alpha$ (learning rate), $\gamma$ (discount factor)}
\State{$\epsilon$ (exploration factor), $\pi_{\epsilon}$ (action sample policy conditioned on $\epsilon$)}
\State{$M$ (maximum number of training episodes), $H$ (horizon length)}

\For{each $\bm{w}$ \textbf{in} $\mathcal{W}^{\eta}$}
    \State {Reset $Env$}
    \State {Initialize a Q table $Q_{\bm{w}}$}
    \For {i\ $\xleftarrow \ 1\ \textbf{to} \  M $}
        \For {t\ $\xleftarrow \ 1\ \textbf{to} \  H $}
            \State {Action sample: $a_{t} \sim \pi_{\epsilon}(s_{t})$ }
            \State {Get reward signal and next state from environment: $\textbf{r}_{t},s_{t+1}=E(a_{t})$}
            \State {Scalarize reward: $r_t = \textbf{r}_{t}\cdot\bm{w}$ }
            \State {Calculate TD error: $\delta_{t}=r_t+\gamma\cdot\ max_{a}Q_{\bm{w}}(s_{t+1},a)-Q_{\bm{w}}({s_t,a_t})$}
            \State {Update Q evaluation: $Q_{\bm{w}}(s_t,a_t) \xleftarrow\ Q_{\bm{w}}(s_t,a_t) + \alpha\cdot\delta_{t}$}
        \EndFor
    \EndFor
    \State{Save $Q_{\bm{w}}$}
\EndFor
\end{algorithmic}
\end{algorithm}

The DWMOTQ algorithm is an effective approach for solving low-dimensional preference and discrete state space problems. It demonstrates success in environments like Deep Sea Treasure where the preference space is only two-dimensional and the state space is one-dimensional, i.e., the position. In terms of convergence speed in such simple environments, DWMOTQ outperforms DWMODQN which sometimes even fails to converge due to the high level of discreteness of the environment.

Nonetheless, in scenarios characterized by image-based state representations and an expanded preference space dimensionality, the computational feasibility of the DWMOTQ algorithm is compromised. As a solution, we use the DWMODQN algorithm, which serves as an alternative strategy to address this challenge.

\subsubsection{Dynamic Weights Multi-objective Deep Q Network}
We adopted the DWMODQN algorithm as an alternative method for approximating the Q function conditioned on the preference weight vector. The DWMODQN algorithm is built upon the deep Q-network (DQN) \cite{mnih2015human} by integrating dynamic weights across multiple objectives \cite{kallstrom2019tunable}.

At the beginning of each episode, the DWMODQN agent samples a preference weight vector $\bm{w}$ as part of its state input and is trained on a high-granularity partitioned discrete subset $\mathcal{W}^{\eta}$ of the preference space. The DWMODQN can learn a single policy that can handle dynamic preference weights without retraining. The DWMODQN algorithm is outlined in Algorithm \ref{alg: Tunable DQN}.
\begin{algorithm}[ht]\small
\caption{DWMORL Algorithm}
\label{alg: Tunable DQN}
\begin{algorithmic}
\State {\textbf{Initialize:} $Env$ (environment), $\alpha$ (learning rate), $\gamma$ (discount factor)}
\State{$\theta$(Q-network), $\theta^{-}$(target Q-network), $\epsilon$ (exploration factor)} 
\State{$\pi_{\epsilon, \theta, \bm{w}}$ (action sample policy conditioned on $\epsilon$, $\theta$ and preference weight $\bm{w}$)}
\State{$M$ (maximum number of training episodes), $H$ (horizon length)}
\State{$\mathcal{D}$ (replay memory), $N_{min}$ (minimum filled capacity of $\mathcal{D}$ to trigger training)}
\For {i\ $\xleftarrow \ 1\ \textbf{to} \  M $}
    \State {Preference weight vector sample $\bm{w} \sim \mathcal{W}^{\eta}$}
    \For {t\ $\xleftarrow \ 1\ \textbf{to} \  H $}
            \State {Action sample: $a_{t} \sim \pi_{\epsilon, \theta, \bm{w}}(s_{t})$ }
            \State {Get reward signal and next state from environment: $\textbf{r}_{t},s_{t+1}=E(a_{t})$}
            \State{Scalarize reward: $r_t = \textbf{r}_{t}\cdot\bm{w}$}
            \State {Add $(s_{t},a_{t},s_{t+1},r_{t},\bm{w})$ to $\mathcal{D}$ }
    \EndFor
    \If{i$>N_{min}$}
        \State {Sample experience batch from $\mathcal{D}$}
        \State Calculate TD error: $\delta_{t}=r_{t} +\gamma\cdot\max_{a}Q_{\theta^{-}}(s_{t+1},a,\bm{w}) - Q_{\theta}(s_{t},a_{t},\bm{w})$ 
        \State Update DQN weights: $Q_{\theta}(s_{t},a_{t},\bm{w}) \leftarrow Q_{\theta}(s_{t},a_{t},\bm{w}) + \alpha\cdot\delta_{t}$
    \EndIf
\EndFor
\end{algorithmic}
\end{algorithm}
\subsection{Truncated Soft Q Action Sampling}
We retained the training process using a deterministic Q estimator of the DWMORL agent and proposed a stochastic sub-optimal action sampling method for demonstration generation, i.e. truncated soft Q action sampling from the trained Q estimators. This is similar to the idea of using an energy-based policy method \cite{haarnoja2017reinforcement, haarnoja2018soft}. However, unlike energy-based policy methods that utilize entropy to improve exploration efficiency and policy robustness, truncated soft Q action sampling aims to generate sub-optimal demonstrations intentionally.

The sub-optimal demonstration assumption has been pointed out in literature \cite{chen2021learning}, i.e. ``The assumption where the user can always provide at least stochastically optimal demonstrations, does not hold in most real-life scenarios.” The method proposed in the referenced work seeks to enhance the efficiency of robotic policy learning by amalgamating sub-optimal demonstrations. Their approach diverges significantly from ours, as our primary objective is to deduce a numerical, vectorized preference factor from the user’s demonstrations, aiming to capture and quantify the underlying preferences.

Furthermore, the focus of the Chen et al. work does not encompass MORL, which is central to our investigation. Our work is specifically tailored to environments characterized by multiple objectives, where understanding and integrating user preferences is crucial for developing policies that accurately reflect the diverse and often competing goals inherent to such settings.

Once fully trained, Q estimators, such as a Q table or a DQN, possess the capability to precisely evaluate state-action pairs. Although in most literature the trained Q estimator simply selects the optimal action, we argue that the full capacity of Q estimators is often underestimated if they are only used for this purpose. Once Q estimators are trained, they can provide a fair comparison between actions, which may include sub-optimal solutions that are also worth considering. The DWMORL agent is trained using $\epsilon$-greedy and this truncated soft Q action sampling is only used during demonstration generation.

\begin{algorithm}[ht]\small
\caption{Truncated Soft Q Action Sampling}
\label{alg: Truncated Soft Q Action Sampling}
\begin{algorithmic}
\State{\textbf{Initialize:}}
\State {Temperature of the Boltzmann distribution of action: $\beta$}
\State{\textbf{Sample action:}}
\State {Estimate Q vector $\bm{Q}(s_{t},\cdot)$ for state $s_{t}$}
\State {Set the last $\lvert \mathcal{A} \rvert-N$ values of $\bm{Q}(s_{t},\cdot)$ as $-\infty$ and get truncated Q vector $\bm{Q}^{+}(s_{t},\cdot)$}
\State {Calculate the probability of action sampling from $\bm{Q}^{+}(s_{t},\cdot)$:}
\State {$\pi_{soft}^{+}(a_{i}\lvert s_{t})=p(a_{i}\lvert s_{t})=\frac{e^{-\beta\cdot Q^{+}(s_{t},a_{i})}}{\sum_{j=0}^{\lvert \mathcal{A} \rvert}e^{-\beta\cdot Q^{+}(s_{t},a_{j})}}$} 
\State{Sample an action from the truncated soft policy $a_{t}\sim \pi_{soft}^{+}(s_{t})$}
\end{algorithmic}
\end{algorithm}

The truncated Q vector is mapped to the probability space, i.e. a Boltzmann distribution. The temperature parameter determines the gap between the probabilities. $N$ actions is retained for sampling from the action space $\mathcal{A}$. While the truncated soft policy $\pi_{soft}^{+}$ is not as guaranteed to be an optimal policy $\pi^{*}$, it is in most cases a good policy as the truncation eliminates most adverse actions while approximately maintaining the distribution of the quality of actions. By generating sub-optimal demonstrations, the training dataset is augmented. This increases the robustness of the inferring model.

\begin{figure}[ht]
    \centering
    \includegraphics[width=10cm]{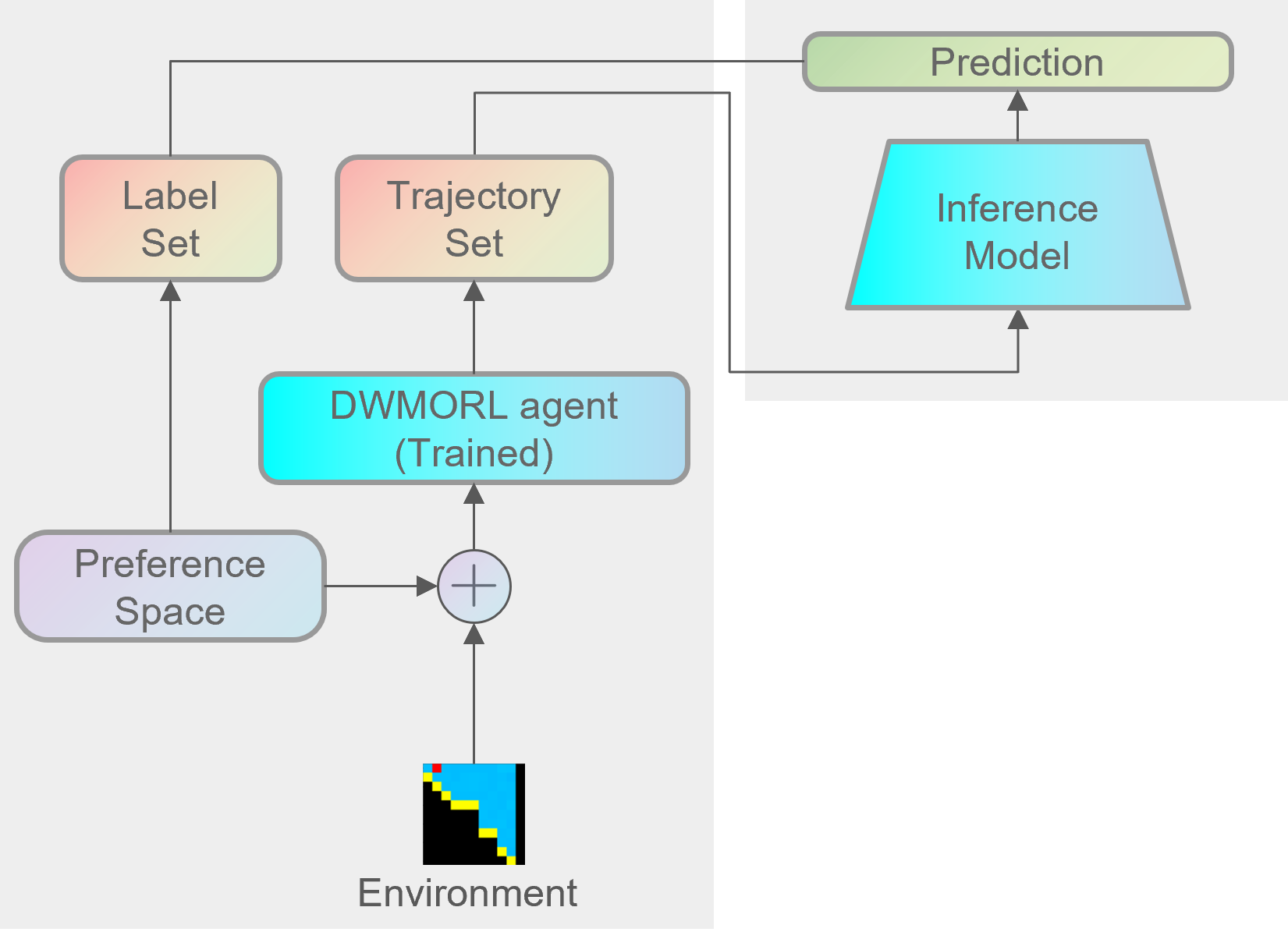}
    \caption{Train the DWPI model}
    \label{fig: Preference Elicitator Training}
\end{figure}

In our experiment, we utilize a combination of optimal and sub-optimal demonstrations to train the DWPI model. Specifically, we make 50\% of the demonstration set optimal demonstrations and the remaining 50\% sub-optimal demonstrations. To further augment the training data, we repeat each demonstration 50 times. The demonstration generation process and the training process of the DWPI model are illustrated in Figure \ref{fig: Preference Elicitator Training}.

During the evaluation, we average the cumulative reward vector of 100 demonstrations for each preference weight and pass them to the inferring model. We ensure that the same ratio of optimal and sub-optimal demonstrations is maintained across all preference vectors.

Notably, in literature, there are works that also try to insert noise into policy to get sub-optimal demonstrations, i.e. D-REX \cite{brown2020better}. However, our method is fundamentally different from D-REX. We deliberately generate sub-optimal demonstrations through truncated soft Q action sampling, aiming to introduce the ``second/third best action" rather than relying on random noise that was adopted in D-REX. This approach is designed to more accurately simulate real-world conditions where demonstrations for PI may not always represent the optimal action. Thus, the application of noise in our research serves a distinct purpose, emphasizing the strategic generation of sub-optimal demonstrations to enrich the dataset for a more accurate and comprehensive PI process.

Another work that mentioned sub-optimal demonstrations is \cite{beliaev2022imitation}. A key distinction between our work and that of Beliaev et al. lies in the setting and application of sub-optimal demonstrations. They focus on extracting valuable insights from sub-optimal demonstrations to enhance policy learning within a singular objective framework. In contrast, our work is situated in a MORL context, where the complexity of balancing competing objectives inherently alters the utility and interpretation of sub-optimal demonstrations. Our approach seeks not to filter or distill behaviors but to infer and quantify underlying user preferences across multiple objectives, a dimension not addressed by the single-objective focus of the prior study.

Moreover, the targets of our work and their work diverge significantly. The aim of our work is not the optimization of policy learning from heterogeneous demonstrators. Instead, our work tackles the challenge of understanding and modeling the preference vectors that drive demonstrator behavior in multi-objective settings. This involves a fundamentally different application of sub-optimal demonstrations, one that serves to illuminate the diverse priorities and decision-making criteria of users rather than to improve policy performance.

\subsection{DWPI Algorithm}
\label{subsec: Overview of the Proposed Method}
Our methodology distinguishes from previous literature by eliminating the necessity for any queries or comparisons among demonstrations, as well as the need to retrain a MORL agent from scratch on each occasion. The DWPI model can also infer demonstrator’s preference from sub-optimal demonstration.

A trained DWMORL agent can generate the optimal demonstration based on its preference weights, a process referred to as ``from preference factors to demonstrations". In contrast, the reverse process, ``from demonstrations to preference factors", can be approached as a regression problem. The demonstration features can serve as the input and the preference weights act as the output labels. This regression model can be implemented with a supervised learning paradigm, as detailed in Algorithm \ref{alg: DWPIAlgorithm}. Initially, we iterate the preference weight space and pass weights into the DWMORL agent to generate demonstrations (the demonstration generation phase). Subsequently, we proceed to train the inference model as a regression model with the demonstrations (inference model training phase). The loss function used for training is the mean squared error (MSE).

\begin{algorithm}[ht]\small
\caption{Dynamic Weight Preference Inference Training Algorithm}
\label{alg: DWPIAlgorithm}
\begin{algorithmic}
\State {\textbf{Initialize:}}
\State {PI model $\phi$, DWMORL agent $\pi$}
\State {\textbf{Demonstration Generation:}}
\For {$\bm{w}_{i}$ in $\mathcal{W}^{\eta}$}
    \State {Sample demonstrations by the DWMORL agent: $\tau_{\bm{w}_{i}} \sim \pi$}
    \State {Add $\tau_{\bm{w}_{i}}$ to training set, add $\bm{w}_{i}$ to label set}
\EndFor
\State {\textbf{Inference Model Training:}}
\While{model $\phi$ not converge}
    \State {Sample $n$ entries from training set, calculate loss $\mathcal{L}=\frac{1}{n}\sum_{i=1}^n(\bm{w}_{i}-\phi_{i}(\tau))^2$}
    \State {Update $\phi$ model: $\phi \leftarrow \phi - \alpha \nabla\mathcal{L}$}
\EndWhile
\end{algorithmic}
\end{algorithm}

\subsection{Theoretical Analysis}
\label{subsec: Theoretical Analysis}
In this section, a theoretical analysis is conducted. We first present a proof of correctness, demonstrating that the DWPI algorithm can be effectively implemented using an FNN. Subsequently, we give the DWPI algorithm's computational complexity. Following this we showcase how feasibility of inferring preference weights from optimal demonstrations. Subsequently, we relax the assumption of optimal demonstrations to include sub-optimal ones.

\subsubsection{Correctness Proof for DWPI}
We present the correctness proof of using an FNN to do PI tasks from demonstrations.
Consider a FNN composed of an input layer, several hidden layers, and an output layer. The network aims to approximate the mapping function: $f:\bm{R}^{n}\xrightarrow{}\mathcal{W}$, all elements on $\mathcal{W}$ are non-negative and sum to 1 \cite{hayes2022practical} and $\bm{R}^{n}$ is the continuous reward space.\\
\textit{Universal Approximation Theorem \cite{cybenko1989approximation, hornik1991approximation} states that, given any continuous function on a compact set and any error greater than zero, there exists a sufficiently large FNN that can approximate this function within an error less than some threshold over the compact set.}\\
In the context of the DWPI algorithm, the continuous function in question maps continuous reward vectors to the preference simplex space. The Universal Approximation Theorem ensures that there exists a neural network structure capable of approximating such a mapping function. This establishes the validity of using an FNN to map demonstrations to preferences, providing a theoretical foundation for DWPI.

\subsubsection{Complexity Analysis}
In this part, we present the computational complexity analysis for Algorithm \ref{alg: DWPIAlgorithm}.
The algorithm is separated into two primary phases: data collection and training. During the data collection phase, the computational complexity is directly proportional to the granularity $\eta$, which denotes the level of discretization applied to the preference simplex. For each discretized element of the preference simplex, the agent is required to execute its policy for an episode. The horizon of the episode and the number of episoded the MORL agent is trained, denoted by 
$H$ and $M$ emerge as critical factors influencing the overall complexity. Consequently, the computational complexity of the data collection phase is expressed in terms of $\mathcal{O}(\eta HM)$, indicating that the time complexity scales with both the granularity of the preference simplex discretization and the length of the episode.

In the training phase, the computational complexity is influenced by the architecture of the FNN utilized. We assume this FNN is full-connected and has $L$ layers, and the $l$th layer has $n_{l}$ neurons. On the $l$th layer, the computational complexity is $\mathcal{O}(n_{l}n_{l-1})$, therefore the whole network feed-forward computational complexity is $\mathcal{O}(\sum_{l=1}^{L}n_{l}n_{l-1})$, the backward propagation has similar complexity to the feed-forward step. Given that there are $n\eta$ demonstrations in the training set, where $a$ is the augmented factor of the training dataset, and the training involves $E$ epochs, the overall complexity for the training process can be expressed as $\mathcal{O}(\eta aE \sum_{l=1}^{L}n_{l}n_{l-1})$. The aggregate computational complexity, accounting for both the generation of data and training is represented by $\mathcal{O}(\eta (HM+aE \sum_{l=1}^{L}n_{l}n_{l-1}))$.

\subsubsection{Mapping Relation from Demonstration to Preference}

The term $\pi_{\bm{w}}$ is the DWMORL agent's policy aligned to the preference weight $\bm{w}$.
Assuming $\bm{w}$ is determined by a user independently of the environment $Env$, 
the demonstration sampled from policy $\pi_{\bm{w}}$ is:

\begin{equation}
\label{eqn: revised demonstration sample}
    \tau_{\bm{w}}\sim\pi_{\bm{w}}
\end{equation}
$ \tau_{\bm{w}}$ is the demonstration conditioned on the preference weight vector $\bm{w}$. Equation \ref{eqn: revised demonstration sample} suggests a mapping relation between the demonstration and preference weight vector in environment $Env$.

An RL agent maps the policy to the demonstration, i.e. $\pi\xrightarrow{}\tau$\cite{sutton2018reinforcement,mnih2015human,kallstrom2019tunable}. This mapping relation is extended in MORL settings as the preference weight is mapped to the demonstration, i.e. $\pi_{\bm{w}}\xrightarrow{}\tau_{\bm{w}}$. The mapping relation ($\pi\xrightarrow{}\tau$) is reversible and introduced as an imitation learning (IL) \cite{ho2016generative,osa2018algorithmic} where the policy is imitated from the demonstration, i.e. $\tau\xrightarrow{}\pi$. 
In the MORL setting, this reversible mapping relation is expanded to the relationship between demonstrations and the preference weight, i.e. $\bm{w}$. Our objective is to establish this mapping relation, denoted as $\tau_{\bm{w}} \xrightarrow{} \bm{w}$. The mapping $\pi_{\bm{w}}\xrightarrow{}\tau_{\bm{w}}$ involves only executing the policy under $\bm{w}$ for several episodes. When using a demonstration $\tau_{\bm{w}}$ to infer $\bm{w}$, the resultant singular, precise weight vector may be different from the label, however, it is still correct if it drops in the correct range of the preference space. Though the policy between the ground truth preference and $\bm{w}$ may be slightly different, they can still be deemed as equivalent under the specific circumstance. We term this as policy equivalence. We first give the formal definition of policy equivalence in the MORL setting then we relax the optimal demonstration assumption and give the formal definition of policy $\delta$-equivalence. We begin with the concept of corner weights proposed in \cite{roijers2016multi}.

The definition of policy equivalence has two primary purposes: 1) It supports the statement that an inference is considered correct if it yields a policy that is equivalent to the expected outcome, even if the inferred preference does not precisely match the ground truth. This approach recognizes the validity of inferences that lead to functionally comparable outcomes. 2) In scenarios where inference is based on a sub-optimal demonstration, a direct comparison between the inferred demonstration and the original sub-optimal one could lead to inaccuracies, resulting in a false negative. In such instances, if the demonstrations are found to be policy $\delta$-equivalent, meaning they achieve a level of performance or outcome within a specified $\delta$ margin, the inference process is deemed to have been successful. This criterion allows for a more nuanced assessment of inference accuracy, accommodating the inherent variability and imperfection in demonstration-based inference.

\begin{definition}
\label{def: corner weight} 
    \textit{Corner Weight ( \cite{roijers2016multi,alegre2023sample})}
    
    \textit{Let $\mathcal{V}=\{\bm{v}^{\pi_{i}}\}_{i}^{n}$ be a set of multi-objective value functions for n policies. Corner weights $\bm{w}_{c}$, are weight vectors located at the vertices of a polyhedron $P$
    \begin{equation}
    P = \{x \in \mathbb{R}^{d+1} \lvert \bm{V}^{+}\bm{x} \leq 0, \sum_{i} w_{i} = 1, w_{i} \geq 0, \forall i\}
\end{equation}
where $\bm{V}^{+}$ is the augmented matrix obtained by appending a column vector of -1's to the matrix $\bm{V}$. Each vector $\bm{x}=(w_{1},w_{2},...,w_{d},v_{\bm{w}})$ in $P$ contains a weight vector $\bm{w}$ and its corresponding value $v_{\bm{w}}$.} 
\end{definition}

The corner weights represent the inner boundaries of the preference space and partition the preference weight space $\mathcal{W}$ into disjoint subsets $\{\mathcal{W}_{i}\}^{\lvert\bm{w}_{c}\rvert}$. When traversing those corner weights the optimal policy $max_{\pi\in\Pi}v_{\bm{w}}^{\pi}$ changes. If a policy is optimal for any one of the weight vectors in the corresponding subset, it is also optimal for all other weights within that subset. 

In the context of MORL, the agent needs to balance between multiple objectives. When using linear preference weight inference, the cumulative reward vector provides a direct determination of the equivalence of two policies. In Definition \ref{def: 4.2}, we give the general concept of policy equivalence.

\begin{definition}
\label{def: 4.2}
\textit{Policy Equivalence}

\textit{$\{\tau_{i}\}^{\pi_{1}}$ and $\{\tau_{i}\}^{\pi_{2}}$}\textit{are demonstration sets from two policies $\pi_{1}$ and $\pi_{2}$. The metric $d$ is used to measure the similarity\footnote{For example, one can employ the MSE between the cumulative rewards of demonstrations, or other metrics capable of quantifying the differences between two demonstrations, as a means of assessment.} of demonstrations. If the two demonstration sets result in the same performance measured by metric $d$:}
\begin{equation}
d(\{\tau_{i}\}^{\pi_{1}},\{\tau_{i}\}^{\pi_{2}})=0
\end{equation}
\textit{$\pi_{1}$ and $\pi_{2}$ are equivalent under metric $d$, noted as:}
\begin{equation}
    \pi_{1}\equiv_{d}\pi_{2}
\end{equation}

\end{definition}
With Definitions \ref{def: corner weight} and \ref{def: 4.2}, we introduce the definition of \textit{CCS regional policy set} to refer to the policy set that is optimal for $\{\mathcal{W}_{i}\}^{\lvert\bm{w}_{c}\rvert}$.

\begin{definition}
\label{def: CCS regional policy set} 
    \textit{CCS Regional Policy Set}

   \textit{For a corner weight $\bm{w}_\mathrm{c}$, all equivalent policies that maximize the scalarized value under $\bm{w}_\mathrm{c}$ constructs the CCS regional policy set $\Pi_{\bm{w}_\mathrm{c}}\subseteq \Pi_{CCS}$. }
\end{definition}


According to Definition \ref{def: corner weight}, $\bm{w}_\mathrm{c}$ determines a region within which various weight vectors are considered equivalent in terms of their ability to generate a policy. Based on Definition \ref{def: CCS regional policy set}, for any weight vector $\bm{w}_{j}$ in a subset bordered by $\bm{w}_{c}$, the policy $\pi_{\bm{w}_{j}}$ are equivalent to the policy $\pi_{\bm{w}_{c}}$ (Definition \ref{def: 4.2}, taking cumulative reward feedback as the equivalence metric $d$). Therefore the inference $\bm{w}_{j}$ is deemed correct as long as the ground truth weight also resides within this range.

\subsubsection{Optimal Demonstration Relaxation}

Note that the DWMORL policy $\pi$ in Algorithm \ref{alg: DWPIAlgorithm} is not necessarily optimal. The inference can still be accurate even if sub-optimal demonstrations are sampled by $\pi$. We now relax the optimal demonstration assumption. Using Definition \ref{def: 4.2}, we define the concept of policy $\delta$-equivalence. 

\begin{definition}
    \label{def: delta equivalence}
    \textit{Policy $\delta$-Equivalence}
    
    \textit{Let $\{\tau_{i}\}^{\pi_{1}}$ and $\{\tau_{i}\}^{\pi_{2}}$ are demonstration sets from two policies $\pi_{1}$ and $\pi_{2}$, and a metric $d$ is used to measure the similarity of demonstrations. If the two demonstrations have a difference less than or equal to a certain threshold $\delta$ under metric $d$:
\begin{equation}
    d(\{\tau_{i}\}^{\pi_{1}},\{\tau_{i}\}^{\pi_{2}})\leq\delta
\end{equation}
$\pi_{1}$ and $\pi_{2}$ are considered equivalent under metric $d$. 
\begin{equation}
    \pi_{1}\equiv_{d_{\delta}}\pi_{2}
\end{equation}}
\end{definition}

The predetermined value of 
$\delta$ should be sufficiently small to ensure that policies associated with distinct preference regions, denoted by $\bm{w}_{c}$, are clearly distinguishable. Specifically, $\delta$ must be constrained to prevent the performance degradation of a sub-optimal policy from being considered to be a policy from another preference region. By using this $\delta$ threshold, when two preference weights—corresponding to a sub-optimal and an optimal policy, respectively—result in $\delta$-equivalence, the preferences are considered to be effectively identical. This criterion aims to maintain a delicate balance: it must be tight enough to preserve the integrity of distinct preference delineations, yet flexible enough to accommodate minor discrepancies that do not materially affect the overall policy efficacy.

We have established a bidirectional mapping between optimal demonstrations and preference weights and elucidated the methodology for validating the correctness of inferences made by the DWPI model when provided with either optimal or sub-optimal demonstrations. This framework not only facilitates the understanding of how preferences can be accurately deduced from demonstrations but also outlines the criteria for assessing the fidelity of these inferences under different demonstration quality levels. This comprehensive approach ensures that the DWPI model's inferences are rigorously evaluated, enhancing its applicability and effectiveness in diverse scenarios where demonstration quality may vary.

\section{Experiment Setting}
\label{sec:Experiment}
In this section, we provide our experimental settings, the baseline algorithms we used, the metrics of evaluation, and the sensitivity analysis setting. The preference weight vectors in all three simulations are normalized during training, the original weights is presented only to keep consistency with the literature.

\subsection{Experiment Environment}
\label{subsec: Experiment Environment}
\begin{figure}[ht]
        \centering
        \includegraphics[width=8.5cm]{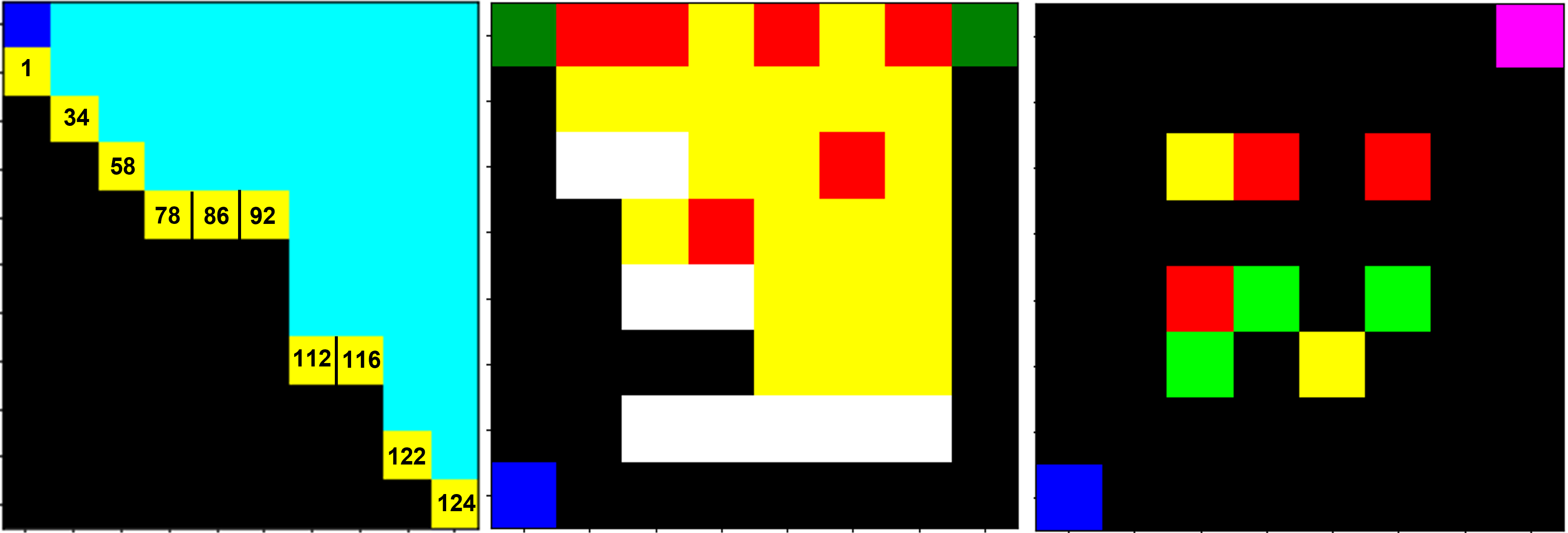}
        \caption{CDST Environment (left): Agent in blue, treasures in yellow with numbers, walkable grids in light blue, unwalkable grids in black. Traffic Environment (middle): Agent in blue, item to collect in green, cars in red, roads in yellow, and walls in white. Item Gathering Environment (right): Agent in blue, fixed-preference agent in pink, three categories collectable items in green, red and yellow. A fixed number of each category of items is randomly placed in the environment at the start of each episode.}
        \label{fig:DST Environment}
\end{figure}
\subsubsection{Convex Deep Sea Treasure}

CDST \cite{mannion2017policy} is a variant of the Deep Sea Treasure (DST) environment \cite{vamplew2011empirical}, where the globally convex Pareto front is known. The state is the current agent position, while the action space consists of four movements: up, down, left, or right. Each episode begins with the agent positioned in the left top corner and ends when the agent reaches any of the yellow treasure grids. There are 10 different treasures, each corresponding to a distinct policy. The agent receives a reward vector comprising two elements: the first element is the time penalty of -1 per time step until the episode terminates, and the second element is the treasure reward. Balancing between the time penalty and treasure reward requires a 2-dimensional preference vector. We implement the DWMORL agent using DWMOTQ, i.e. Algorithm \ref{alg: Q Learning with tunable training}.

\subsubsection{Traffic}
The Traffic environment \cite{kallstrom2019tunable} has a collectible item on each of the upper corners. The yellow area is the road part where the red cars move. The car moves vertically in a random direction and it reverses the direction when hit by the wall or the edge of the frame. The agent, starting from the left bottom corner, is not supposed to step on the road. If the agent steps on the road, it risks being hit by a car.

It must make a trade-off between obeying the traffic rules, maintaining traffic safety, and the time it takes to collect items. Each item is worth reward 1, stepping on the road causes reward -1, being hit by a car causes reward -1, hitting on the wall or the edge of the frame causes reward -1, and the time penalty is -1. In the work of Kallstrom et al. \cite{kallstrom2019tunable}, four different simulation scenarios are given, leading to four typical behavior patterns. We retain the original preference setting from the literature to keep consistency. The elements of the preference vector are ordered as [\textit{steps, item collection, break traffic rules, collisions, wall hitting}]. The DWMORL agent is implemented with Algorithm \ref{alg: Tunable DQN}. The scenarios are shown below.\\
\textbf{Always Safe:} [1, 50, 10, 50, 1]\\ 
The agent tries its best to avoid illegal behavior and collisions. It takes the longest but safest path to collect the two items.\\
\textbf{Always Fast:} [10, 50, 10, 10, 1]\\ 
The agent tries to collect the two items as fast as possible. It thinks a detour is costly and therefore does not care about the risk of collision and breaking traffic rules by taking a short path.\\
\textbf{Fast and Safe:} [5, 50, 0, 50, 1]\\
The agent cares about time consumption but also the safety of the path. It is encouraged to break traffic rules and walk on the yellow path to spend less time.\\
\textbf{Slow and Safe:} [1, 50, 0, 50, 1]\\
The agent will walk on the road if there is a low risk of being hit by cars. It would rather take a longer path if the traffic condition is not ideal.

\subsubsection{Item Gathering}
The Item Gathering environment \cite{kallstrom2019tunable} features three categories of collectibles in different colors. There is a hard-coded agent with fixed preferences that particularly favors the red item. At the start of each episode, the positions of the items are randomly initialized. The hard-coded agent starts at the top right corner and picks random routes to the red items.

The DWMORL agent exhibits an altruistic preference towards the hard-coded agent's collection of red items, which determines its cooperative or competitive behavior. The DWMORL agent obtains a reward of 1 that is linked to the objective of collecting an item of a particular color. An additional reward of 1 for its altruistic objective if the hard-coded agent collects a red item. The altruistic reward will be reversed based on the DWMORL agent's cooperation or competition behavior that depended on the symbol of the altruistic preference. The DWMORL agent receives a time penalty of -1 at each time step and a wall penalty of -1 upon collision.

Kallstrom et al. \cite{kallstrom2019tunable} examined four simulation scenarios to illustrate distinct behavior patterns of the DWMORL agent. We retain their setting to keep consistency. The preference vector's elements were arranged in the following order: [\textit{time penalty, wall penalty, collection of green items, collection of red items, collection of yellow items, and collection by the hard-coded agent}].\\
\textbf{Competitive: }[1, 5, 10, 20, 10, -20]\\
The DWMORL agent's competitive behavior poses a challenge for the hard-coded agent in collecting red items.\\
\textbf{Cooperative: }[1, 5, 10, 20, 10, 20]\\
While both agents prioritize the collection of red items, the DWMORL agent's behavior is not characterized by vicious competition.\\
\textbf{Fair: }[1, 5, 20, 15, 20, 20]\\
The DWMORL agent prioritizes the collection of yellow and green items and is content with the success of the hard-coded agent.\\
\textbf{Generous: }[1, 5, 20, 0, 20, 20]\\
The DWMORL agent actively ignores and even avoids red items to promote the success of the hard-coded agent.\\

\subsection{Baseline}
\label{subsec:Baseline}
We re-implement the PM method \cite{ikenaga2018inverse}, and the MWAL method \cite{takayama2022multi},  as the two baseline algorithms to compare our DWPI algorithm against. The PM method randomly starts from a preference, trains a RL agent from scratch and compares the resulting cumulative reward with the cumulative reward from the given demonstration. Similar to the PM method, MWAL method infers preferences by comparing the agent's and the given cumulative reward (from demonstration). Instead of randomly searching for preferences as PM, they use Equation \ref{eqn: MWAL} to update the preference element, where $w_{n}$ is the nth element of the preference, $k$ is the number of feature elements, $N$ is the total number of iterations, and $\mu_{n}$ and $\mu_{E_{n}}$ is the nth element of the feature expectation from current RL agent and the expert. 
\begin{equation}
\label{eqn: MWAL}
    w_{n}\xleftarrow{} w_{n}\cdot(1+\sqrt{\frac{2logk}{N}})^{-(\mu_{n}-\mu_{E_{n}})}
\end{equation}

To make fair comparison, we used the same RL algorithm for PM and MWAL as the DWPI method.

\subsection{Metrics}
The research of PI is still at its early stage, yet a widely accepted metric is established. We therefore propose the following metrics to determine the accuracy of PI models.
\subsubsection{Time Efficiency} 
Efficient PI is crucial for practical applications that require rapid decision-making, especially in high-dimensional preference spaces. We evaluate the time consumption of the DWPI method and baselines by calculating the average time required for training phase and inference phase. 

\subsubsection{Inference Accuracy}
\label{subsubsec: Inference Accuracy}
\paragraph{Known Pareto Front}
In CDST, the Pareto front is already known. In this scenario, the inference accuracy can be assessed by traversing through $\mathcal{W}^{\eta}$ and verifying whether the inferred preference weight falls within the correct range. The inference accuracy can then be calculated using the following formula, where $N_{correct}$ is the number of correct inferences:
\begin{equation}
    \mathcal{ACC} = \frac{N_{correct}}{\lvert\mathcal{W}^{\eta}\rvert}
\end{equation}

\paragraph{Unknown Pareto Front}
The direct comparison is useful for evaluating inference accuracy when Pareto front is known. However, in real-life scenarios, a known Pareto front is a rare occurrence and a more frequent case is to do PI task with a unknown Pareto front.

Moreover, in MORL, the policy space often exhibits imbalanced clusters. A slight change in preference may lead to significant policy variations. This hinders the use of value-based comparison methods, e.g. MSE, for accuracy evaluation. In some cases, the ground truth and the inference may be very close to each other, but drop on two different sides of the boundary of two different policies (the corner weight). This indicates a plausible but incorrect inference and impact the evaluation adversely. 

To address these two issues, comparing the cumulative rewards generated from the inferred preference weight with those generated from the true preference weight is a more reasonable evaluation metric of the inference accuracy. This approach is particularly effective in cases where a small preference change can result in a significant policy variation. If MSE is used to compute the difference between the cumulative rewards, we can use the following formula, where $\bm{w}$ is the ground truth and $\hat{\bm{w}}$ is the inference:
\begin{equation}
    \mathcal{ACC} = (\sum_{\tau_{\hat{\bm{w}}}}\bm{r}-\sum_{\tau_{\bm{w}}}\bm{r})^2
\end{equation}
\subsection{Sensitivity Analysis Experiment Setting}
To evaluate the DWPI model when using different types of demonstration, we conducted a sensitivity analysis by assessing performance with different demonstration representations.

\subsubsection{Varying demonstration Representation}
The DWPI model is trained with cumulative rewards (demonstration reward) by default. However, there are other representations of a demonstration, e.g. state demonstration. It is worthwhile to examine the difference in performance while different representations are used in PI.

In this section, we give the experiment settings to evaluate the performance of the DWPI model trained with state demonstrations. As the three simulation environments are all based on a grid world framework where the agent position is finite and discrete. We number the agent position and use a visiting frequency to denote the state demonstration. The state-frequency demonstration are calculated as:
\begin{equation}
    \tau_{s} = \sum_{t=1}^{\lvert\tau_{s}\rvert}\bm{e}_{s}(t)\cdot\bm{I}_{\lvert\mathcal{S}\rvert}^{T}
\end{equation}
where $\bm{I}_{\lvert\mathcal{S}\rvert}$ is an identity matrix with shape $\lvert\mathcal{S}\rvert \times \lvert\mathcal{S}\rvert$, $\bm{e}(s_{t})$ is the one-hot vector that has one in the position corresponding to state $s_{t}$.
The state demonstrations are averaged from 100 demonstrations, similar to the reward demonstrations. The training hyperparameters are the same as the reward demonstration setting, except for the input layer's structure.

\section{Results and Analysis}
\label{sec:Result}
\subsection{Time Efficiency}
To ensure a fair comparison, we terminate the baselines once the error of PI converges. The time consumption of DWPI includes not only the PI process but also the training process of both the DWMORL model and the DWPI model.

\begin{figure}[ht]
        \centering
        \includegraphics[width=8.5cm]{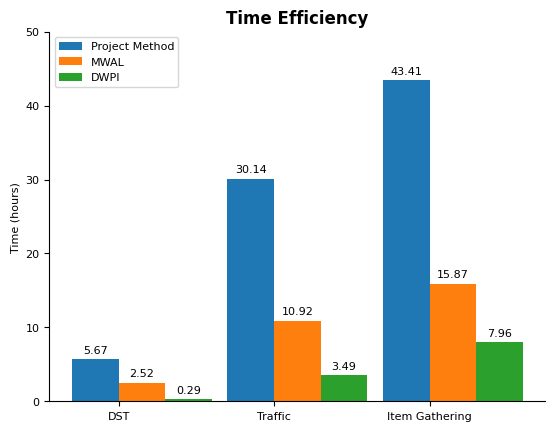}
        \caption{Time Efficiency Comparison}
        \label{fig: Time Efficiency Comparison}
\end{figure}

The results are shown in Figure \ref{fig: Time Efficiency Comparison}. The figure illustrates that the DWPI algorithm demonstrated superior time efficiency compared to the baseline PM and MWAL algorithms.

Specifically, in the CDST simulation, DWPI was 94.83\% faster than PM and 88.35\% faster than MWAL; in the Traffic simulation, DWPI was 88.41\% faster than PM and 68.02\% faster than MWAL; and in the Item Gathering simulation, DWPI was 81.67\% faster than PM and 49.85\% faster than MWAL.

The results show that the DWPI algorithm is significantly more time-efficient for PI tasks in these three simulations compared to the baseline algorithms.

\subsection{Inference Accuracy}
\begin{figure}[ht]
        \centering
        \includegraphics[width=8.5cm]{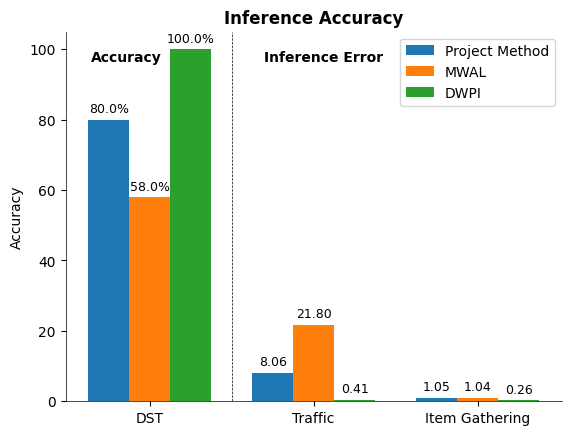}
        \caption{Inference accuracy Comparison}
        \label{fig: Inference accuracy Comparison}
\end{figure}

The overall inference accuracy comparison between the DWPI algorithm and baselines is shown in Figure \ref{fig: Inference accuracy Comparison}.

In CDST simulation, the direct comparison of inference and ground truth is adopted as the Pareto front is already known. Compared with the PM and MWAL algorithms, the DWPI algorithm shows 20\% and 42\% higher inference accuracy, respectively. Additionally, our approach achieves perfect accuracy (100\%) in inference tasks for demonstrations with a 50\% stochastic ratio. Further information regarding the results of the CDST simulation can be found in Section \ref{subsubsec: DST environment}.

Inference error is used for accuracy evaluation in the Traffic environment and the Item Gathering environment as their Pareto fronts are not known. The inference is operated on the resulted demonstrations of the four typical preference entries, the stochastic ratio is also 50\% to evaluate the algorithms' robustness.  In the Traffic simulation, DWPI achieves 94.91\% lower inference error than the PM, and 98.12\% lower inference error than the MWAL method. In the Item Gathering simulation, DWPI achieves 75.24\% lower inference error than the PM and 75\% lower inference error than the MWAL method. 

It is worth noting that the MWAL algorithm consistently demonstrated the lowest inference accuracy across both CDST and Traffic simulations and similar performance as PM in Item Gathering simulation. This is attributed to the fact that the MWAL algorithm tends to perform poorly in situations where there are sub-optimal demonstrations and a stochastic environment. In contrast, the PM, while is not time-efficient, exhibits more robust performance than MWAL due to its use of a ``sample-and-try-out" pattern. This enables the PM to make relatively good inferences even in situations with sub-optimal demonstrations. Our DWPI algorithm surpasses both MWAL and PM in terms of time efficiency and inference accuracy. In the upcoming sections, we will outline the specific details and outcomes of each simulation.

\subsubsection{DST environment}
\label{subsubsec: DST environment}
\begin{figure}[ht]
        \centering
        \includegraphics[width=12cm]{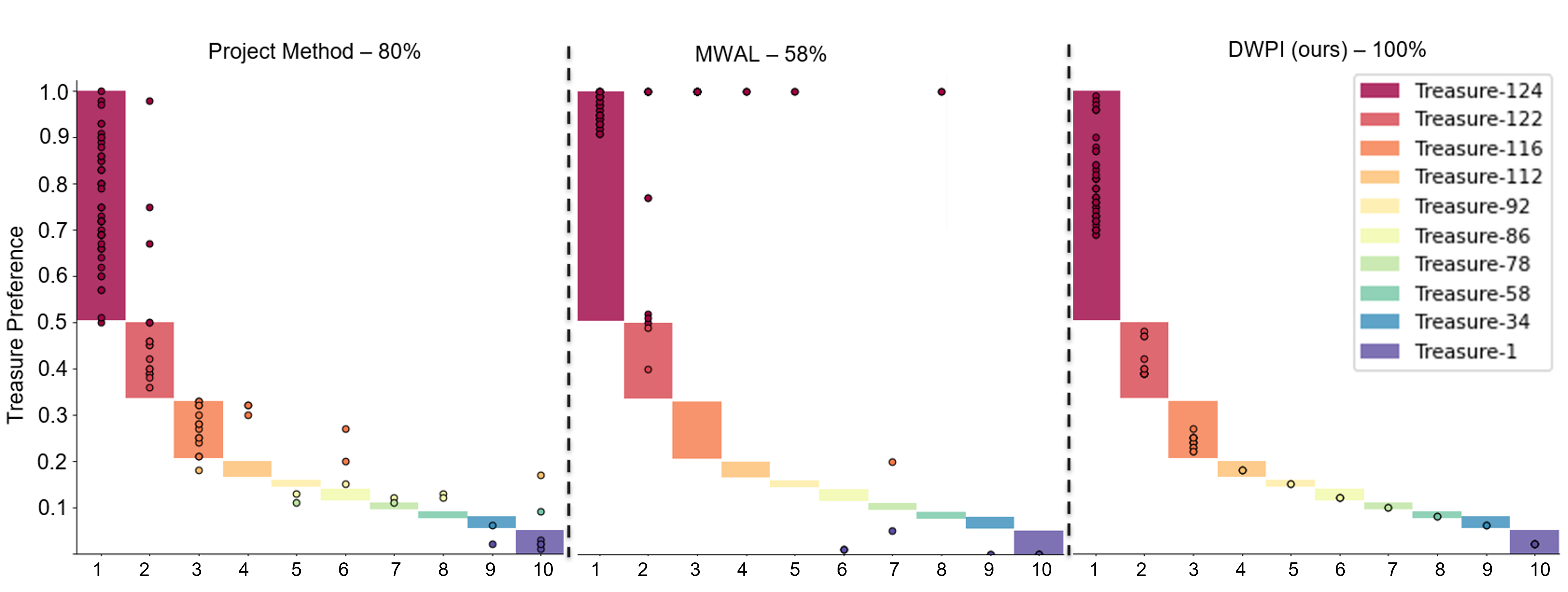}
        \caption{Inference Result CDST - Stochastic demonstration Ratio - 50\%}
        \label{fig: DST result 0.5}
\end{figure}

In the CDST simulation, we compare the DWPI algorithm to baseline algorithms according on their inference accuracy. A more detailed comparison is presented in Figure \ref{fig: DST result 0.5}. Both PM and MWAL perform well when demonstrations are going to Treasures 124 and 122 because they are belong to the two largest policy regions. However, the performance of MWAL and PM deteriorates within policy regions mapped by the middle area of preference space. This is because those preference regions are more subtle and small, and even a small change in preference weight may incur a totally different policy. In the case of stochastic sub-optimal demonstration, while PM has a chance to infer a near-correct preference vector by sample and try-out, MWAL suffers from the difference between the sub-optimal demonstration and the optimal demonstration. When the preference range expands, both MWAL and PM exhibit improved inference accuracy, as demonstrated in the Treasure-1 category, represented by the purple region. Over the two baselines, our DWPI outstands with 100\% accuracy in CDST simulation which is shown in the right-most Figure.

\subsubsection{Traffic environment}
\label{subsubsec: Traffic environment}
\begin{figure}
  \centering
  \begin{subfigure}[b]{1\textwidth}
    \includegraphics[width=\textwidth]{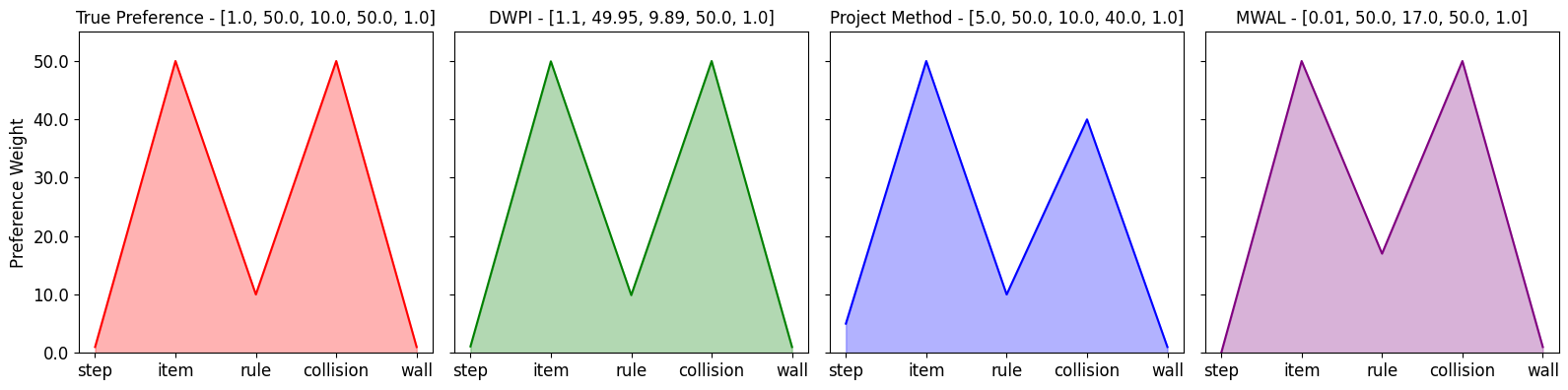}
    \caption{Always Safe}
    \label{fig:sub2}
  \end{subfigure}
  \hfill
  \begin{subfigure}[b]{1\textwidth}
    \includegraphics[width=\textwidth]{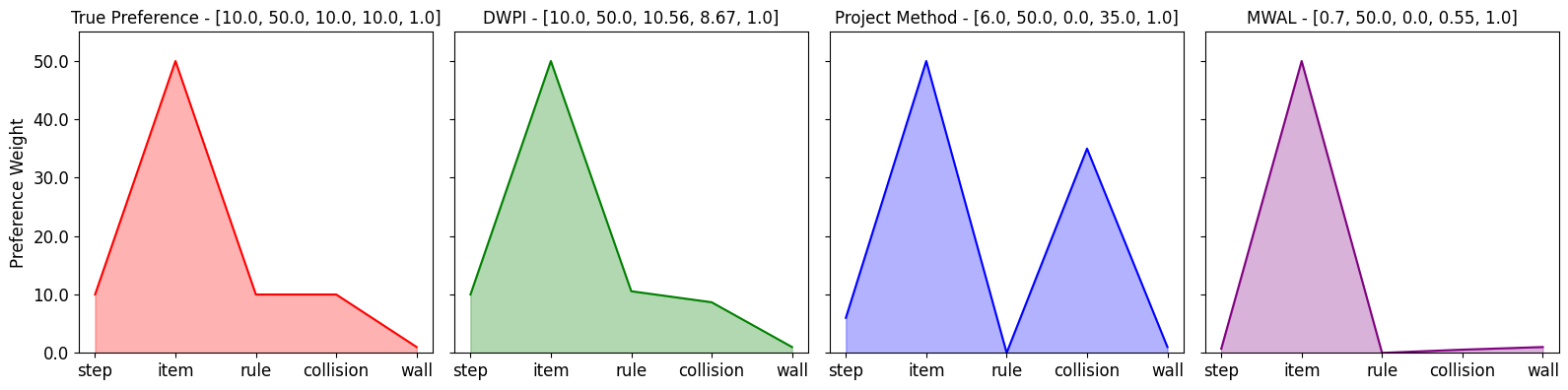}
    \caption{Always Fast}
    \label{fig:sub1}
  \end{subfigure}
  \hfill
  \begin{subfigure}[b]{1\textwidth}
    \includegraphics[width=\textwidth]{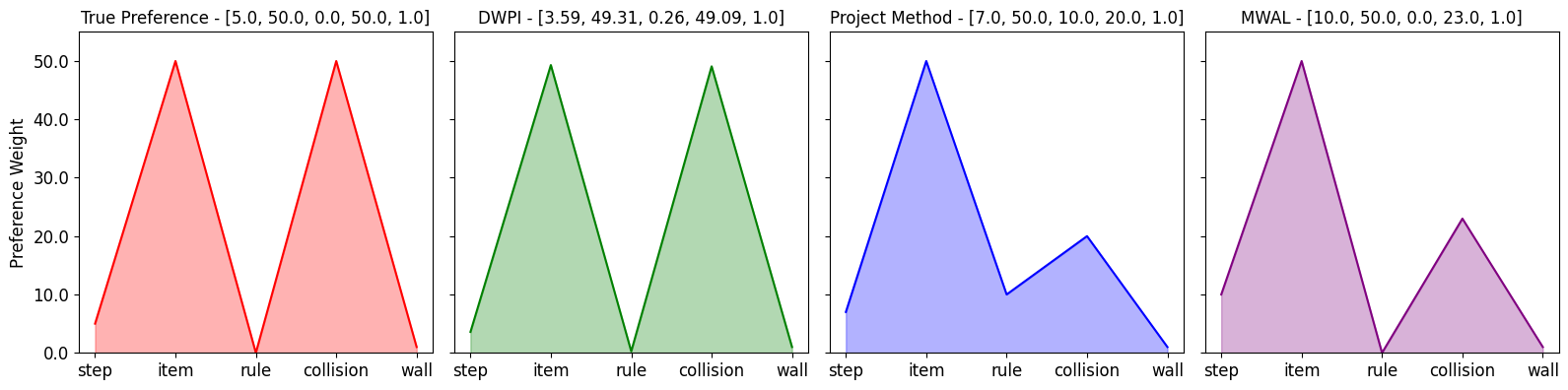}
    \caption{Fast Safe}
    \label{fig:sub4}
  \end{subfigure}
  \hfill
  \begin{subfigure}[b]{1\textwidth}
    \includegraphics[width=\textwidth]{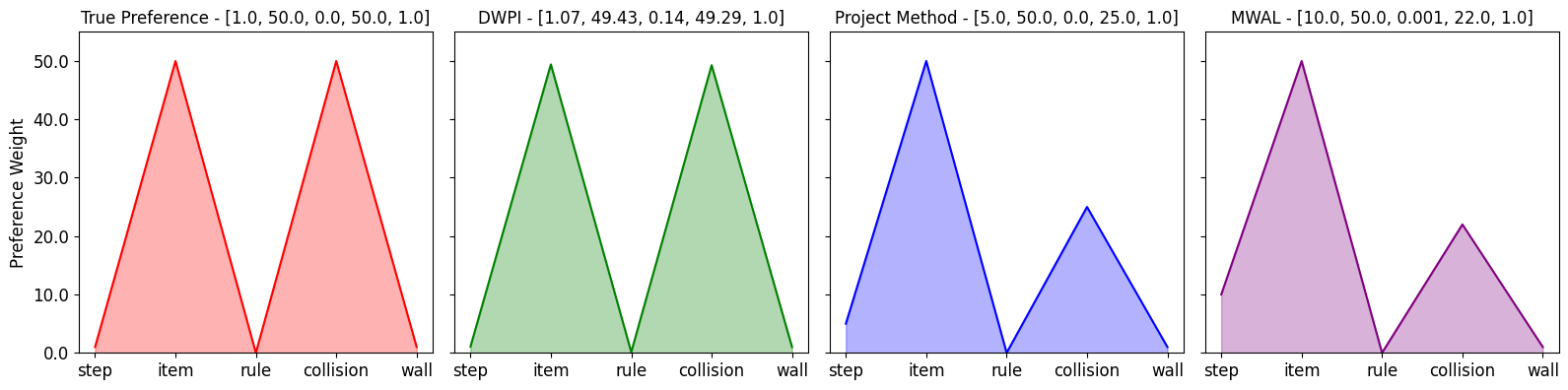}
    \caption{Slow Safe}
    \label{fig:sub3}
  \end{subfigure}
  \caption{Inference Result Traffic - Stochastic demonstration Ratio - 50\%}
  \label{fig: Traffic Inference Result}
\end{figure}

Figure \ref{fig: Traffic Inference Result} illustrates the outcome of PI in Traffic simulation. We assessed the inference accuracy in four common scenarios: always safe, always fast, fast safe, and slow safe. The red color represents the ground truth of the preference weight, while the green, blue, and purple colors represent the inference of DWPI, the PM, and MWAL, respectively. It is worth noting that the weights are in line with the experiment setting used by Kallstrom et al. \cite{kallstrom2019tunable}, and normalization during training or inference would not affect their generality. The y-axis is the preference weight, and the x-axis labels are the objectives, i.e. [\textit{steps, item collection, break traffic rules, collisions, wall hitting}]. The numeric inference results are put in the title of each subplot. 

Based on the results depicted in Figure \ref{fig: Traffic Inference Result}, both the MWAL and PM approaches are able to capture the general shape of the true preference, however, there are some deviations to decrease the performance. In the ``always fast" case, the MWAL method reduces too much the preference for avoiding collision and hitting the wall, while the PM shows two peaks in the preference distribution, despite the two preference weights actually being a downslope. The MWAL method performs well in the deterministic "always safe" case, where the agent picks the safest road and avoids breaking rules or being hit by a car in a grid-world environment. However, the MWAL method is brittle in cases with stochasticity, such as ``always fast ", ``fast safe", and ``slow safe". On the other hand, the PM performs reasonably well in most cases and is not very sensitive to stochasticity, but it suffers from time efficiency and performs poorly in the ``always fast" case.

The poor performance of baselines in the ``always fast" case is due to the high stochasticity of the environment when the demonstrator wants to be always fast, and the agent goes on a road with many cars. This introduces significant stochastic noise to the reward demonstration, causing both baselines to almost fail to infer the preference weight. In contrast, the DWPI model, which can be trained by stochastic demonstrations, shows robustness in not only the ``always fast" case but also perfectly infers the preference for the other three cases. As shown in Figure \ref{fig: Traffic Inference Result}, Figure \ref{fig: Time Efficiency Comparison}, and Figure \ref{fig: Inference accuracy Comparison}, our DWPI model can almost perfectly infer the preference weights within the shortest time.

\subsubsection{Item Gathering environment}
\begin{figure}
  \centering
  \begin{subfigure}[b]{1\textwidth}
    \includegraphics[width=\textwidth]{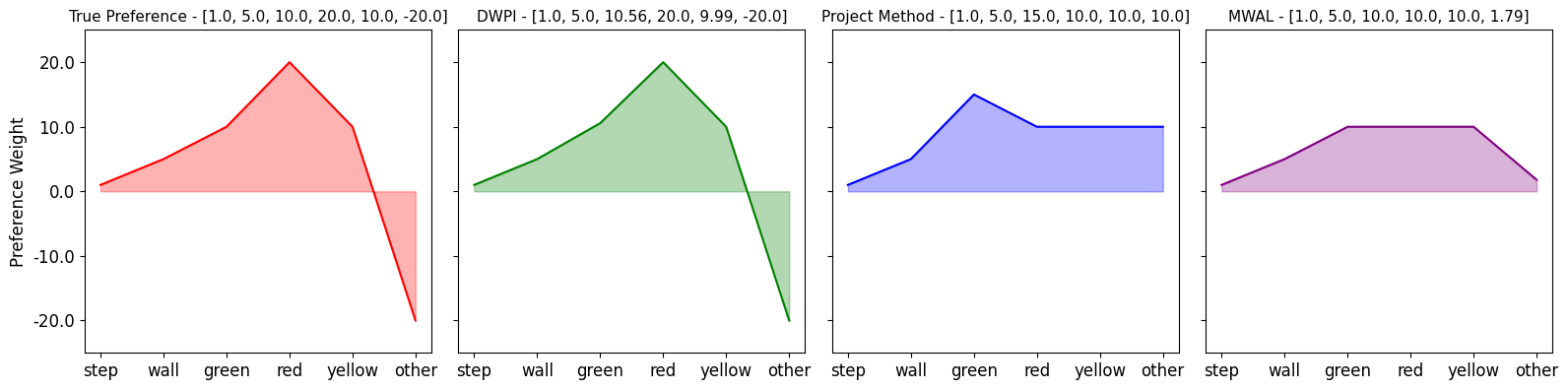}
    \caption{Competitive}
    \label{fig:sub20}
  \end{subfigure}
  \hfill
  \begin{subfigure}[b]{1\textwidth}
    \includegraphics[width=\textwidth]{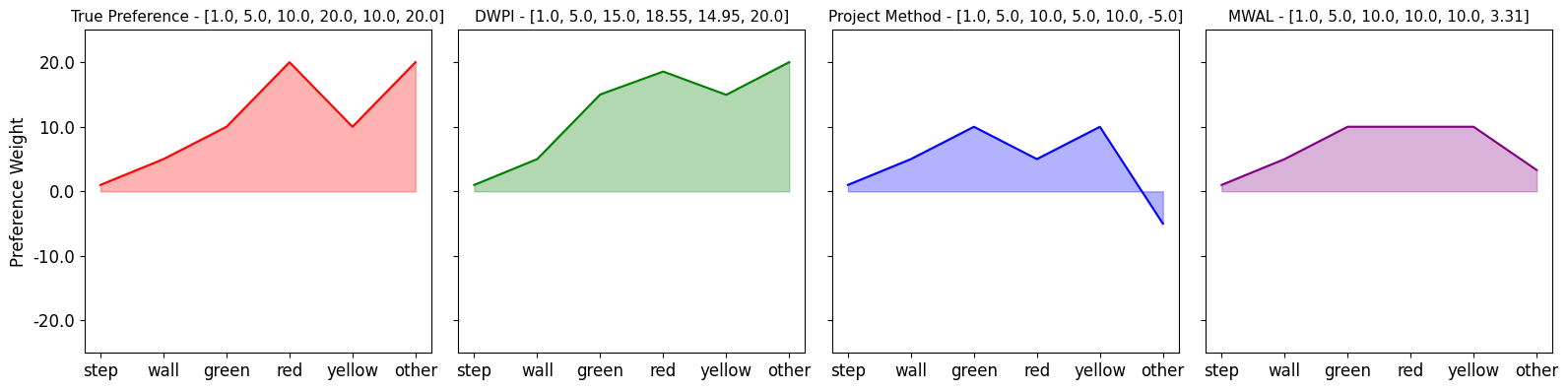}
    \caption{Cooperative}
    \label{fig:sub10}
  \end{subfigure}
  \hfill
  \begin{subfigure}[b]{1\textwidth}
    \includegraphics[width=\textwidth]{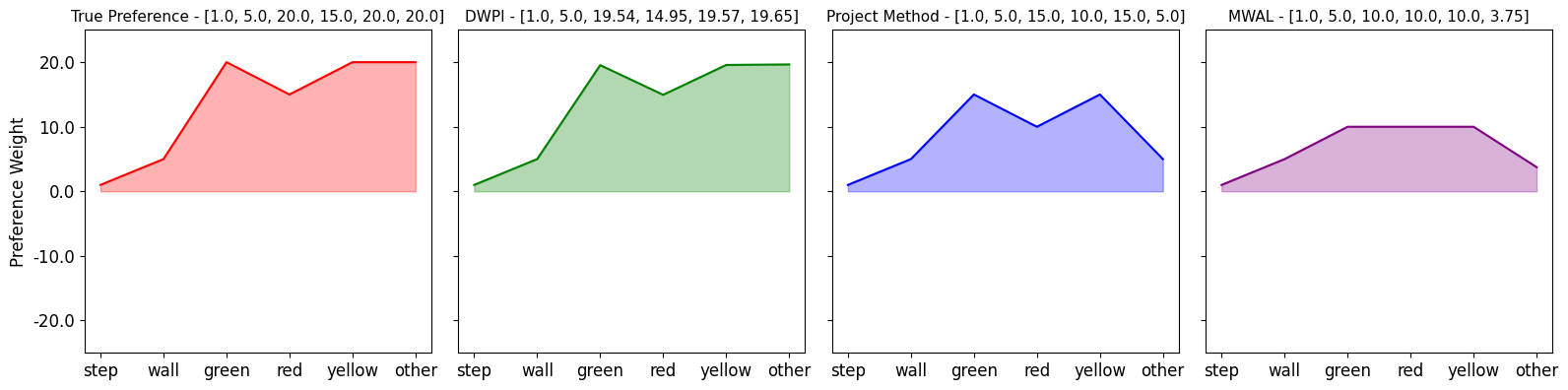}
    \caption{Fair}
    \label{fig:sub30}
  \end{subfigure}
  \hfill
  \begin{subfigure}[b]{1\textwidth}
    \includegraphics[width=\textwidth]{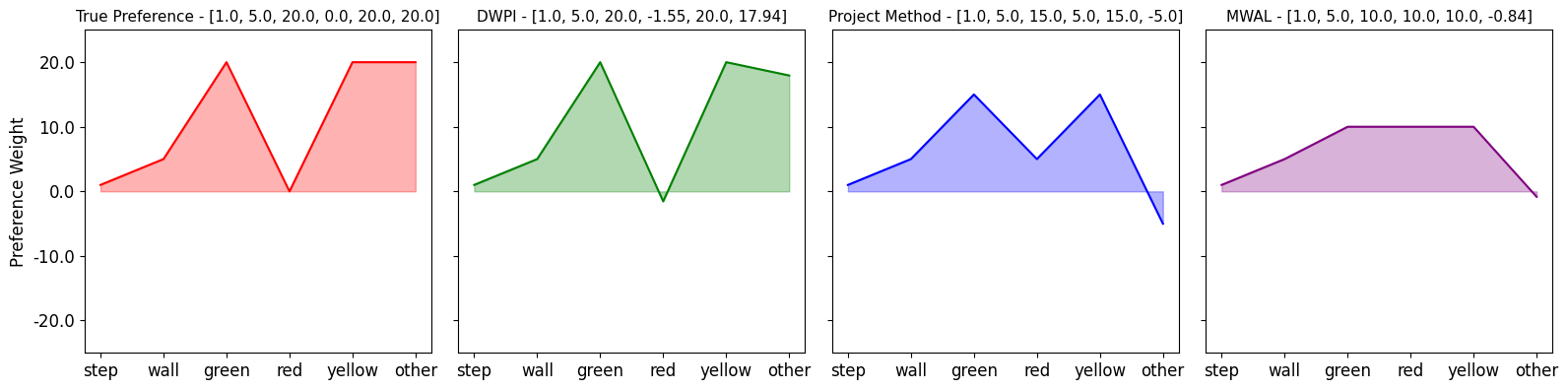}
    \caption{Generous}
    \label{fig:sub03}
  \end{subfigure}
  \caption{Inference Result Item Gathering - Stochastic demonstration Ratio - 50\%}
  \label{fig: Item Gathering Result}
\end{figure}

Figure \ref{fig: Item Gathering Result} showcases the results of the Item Gathering simulation where we evaluate the inference performance of the DWPI, PM, and MWAL for the competitive, cooperative, fair, and generous behavior patterns of the agent. Similar to Figure \ref{fig: Traffic Inference Result}, the ground truth is shown in red while the results from the three methods are shown in green, blue, and purple. The preference weights are consistent with the ones used in literature \cite{kallstrom2019tunable}, and normalization is applied for generality. The x-axis represents the objectives such as time penalty, wall penalty, and item collections. The numeric inferences are displayed in to the subplot title.

The results in Figure \ref{fig: Item Gathering Result} indicate that our DWPI model can capture not only the general shape of the true preference but also demonstrate high accuracy numerically, including negative weights. In contrast, the PM can only capture the rough shape of the ground truth and fails to infer the negative weight for the competitive behavior pattern, and cannot provide distinguishable results for the cooperative and fair behavior patterns. Due to the high stochasticity of the environment where items are randomly dropped at the start of the episode, MWAL cannot give accurate inferences and even converges to incorrect preference weights.

\label{subsubsec: Item Gathering environment}
\subsection{Varying demonstration Representation Sensitivity Analysis}
\begin{figure}[ht]
        \centering
        \includegraphics[width=8.5cm]{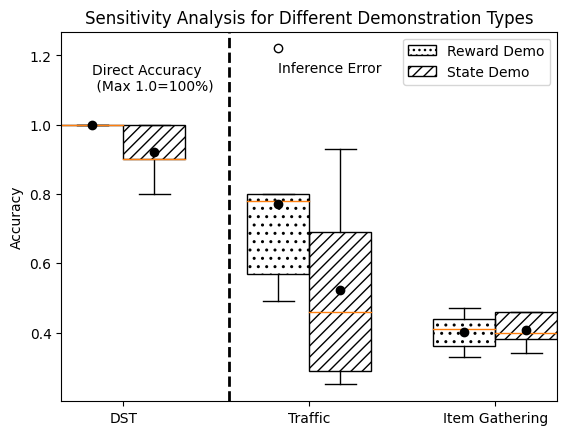}
        \caption{Inference Accuracy Comparison Varying demonstration}
        \label{fig: Inference Accuracy Comparison Varying demonstration}
\end{figure}

In this section, we present the results of sensitivity tests, where we evaluate the inference performance of DWPI for different categories of demonstration. We run each evaluation for 5 times with different random seeds\footnote{Random seeds: 42, 17, 2, 24, 96}. The results of inference from reward demonstration and state demonstration are shown in Figure \ref{fig: Inference Accuracy Comparison Varying demonstration}. Similar to Figure \ref{fig: Inference accuracy Comparison}, the metric of CDST simulation is the direct accuracy while the metric for Traffic simulation and Item Gathering simulation is inference error. 

The inference accuracy from the reward demonstrations is better than the result from state demonstrations in CDST environment. However, in Traffic environment the state demonstrations is more efficient and effective for inference while the variance of state demonstration is slightly larger. In the Traffic simulation, when using reward demonstraion, there is an outlier which should be caused by the high stochasticity of the environment. In the Item Gathering environment, the two representations of demonstration achieve competitive result.

From the difference between the results from the two different types of demonstrations, we have two assumptions. 1) The results obtained from the simulation environments used in our study show that reward demonstration is a more effective descriptor than state demonstration to quantitatively infer the preference. However, as in Traffic and Item Gathering, the state demonstration is more effective to infer the preference that incur less variance of the resulted policy. 2) The inference performance of state demonstration improves when the environment becomes more stochastic. This may because that when the environment becomes more stochastic, even under the same preference, the agent may achieve very different rewards. However, the state sequence is more reliable to link with the preference in this scenario.

\subsection{Discussion}
We evaluated the performance of our proposed method against two baseline methods, PM and MWAL, in three different simulations: DST, Traffic, and Item Gathering. Our proposed DWPI approach demonstrated superior time efficiency and inference accuracy in all three simulations compared to the baseline methods.

In the time efficiency comparison, our DWPI algorithm outperformed the PM and MWAL in all three simulations.
In the inference accuracy comparison, our DWPI algorithm demonstrated higher accuracy in DST, Traffic, and Item Gathering simulations compared to the baseline methods.

We also conducted a sensitivity analysis to evaluate the performance of our DWPI algorithm for different categories of demonstration representation. The results indicate that the reward demonstration is more effective in inferring preference than the state demonstration in all three simulations. The difference between the two representations becomes less pronounced as the dynamic nature of the environment becomes more stochastic, which is an important consideration when applying this method to real-world problems.

\section{Conclusion}
\label{sec:Conclusion}

We introduce a novel approach ``dynamic weight-based preference inference" for PI in multi-objective reinforcement learning. It is implemented based on a dynamic weight multi-objective reinforcement learning agent. The proposed algorithm outperforms existing baseline approaches in terms of accuracy and computational efficiency. It is capable of inferring user preferences even for sub-optimal demonstrations. 
\begin{itemize}
    \item The DWPI algorithm, a time-efficient, robust, query-free, and high-accuracy method for PI task in MORL settings. The trained model simply does a forward pass to infer the preference of a demonstration.
    \item Three distinct environments are used to evaluate the ability of generalization of the DWPI. Each of the three environments features a varying number of objectives.
    \item We provide a proof of correctness for the using a FNN for PI tasks and the complexity analysis of our algorithm. We also introduce a metric to evaluate the accuracy of the inference process.
    \item To deliberately generate sub-optimal demonstrations we use an energy-based model. This is done for augmenting the training dataset.
    \item We conduct a sensitivity analysis of DWPI model to evaluate the performance under different demonstration representations.
\end{itemize} 
We aim to explore real human demonstrations in future work. Furthermore, incorporating representative learning techniques to improve the representation of the demonstration is another promising way to improve this work. Moreover, several avenues for future research are identified, including evaluating the DWPI algorithm in multi-agent environments, extending the algorithm to handle non-linear utility functions, and more metrics to evaluate the accuracy of inference and the application in real-world problems.
\section{Data Availability}
We have released the code of this work on https://github.com/JLu2022/DWPI . 

\section{Conflict of Interest}
This research is funded by an Irish Research Council Government of Ireland Postgraduate Scholarship (GOIPG/2022/2140).
All authors disclosed no relevant relationships.
\bibliographystyle{plainnat}
\bibliography{DWPI}


\end{document}